\documentclass[10pt,twocolumn,letterpaper]{article}
\usepackage[accsupp]{axessibility}  
\usepackage{iccv}
\usepackage{times}
\usepackage{epsfig}
\usepackage{graphicx}
\usepackage{amsmath}
\usepackage{amssymb}
\usepackage{enumitem}
\usepackage{soul}
\usepackage{multirow}

\usepackage{amsmath,amsfonts,bm}

\def\eqref#1{equation~\ref{#1}}

\def\1{\bm{1}}

\def\mO{{\bm{O}}}
\def\mP{{\bm{P}}}

\def\mW{{\bm{W}}}

\def\mZ{{\bm{Z}}}

\DeclareMathAlphabet{\mathsfit}{\encodingdefault}{\sfdefault}{m}{sl}
\SetMathAlphabet{\mathsfit}{bold}{\encodingdefault}{\sfdefault}{bx}{n}

\setlist{nolistsep}
\usepackage{booktabs}
\newcommand{\res}[2]{$#1_{\pm#2}$}

\usepackage[pagebackref=true,breaklinks=true,letterpaper=true,colorlinks,bookmarks=false]{hyperref}

\iccvfinalcopy 

\def\iccvPaperID{6396} 

\ificcvfinal\fi

\begin{document}

\title{FedPerfix: Towards Partial Model Personalization of Vision Transformers in Federated Learning}

\author{Guangyu Sun\textsuperscript{1}, Matias Mendieta\textsuperscript{1}, Jun Luo\textsuperscript{2}, Shandong Wu\textsuperscript{3}, Chen Chen\textsuperscript{1}\\
\\
\textsuperscript{1}Center for Research in Computer Vision, University of Central Florida, USA\\
\textsuperscript{2}Intelligent Systems Program,
University of Pittsburgh, USA\\
\textsuperscript{3}Department of Radiology, Biomedical Informatics,\\ and Bioengineering,
University of Pittsburgh, USA\\
\small{\tt{\{guangyu.sun, matias.mendieta\}@ucf.edu; jul117@pitt.edu;}} \\
\small{\tt{wus3@upmc.edu; chen.chen@crcv.ucf.edu}} 
}

\maketitle
\ificcvfinal\thispagestyle{empty}\fi

\begin{abstract}
Personalized Federated Learning (PFL) represents a promising solution for decentralized learning in heterogeneous data environments. Partial model personalization has been proposed to improve the efficiency of PFL by selectively updating local model parameters instead of aggregating all of them. However, previous work on partial model personalization has mainly focused on Convolutional Neural Networks (CNNs), leaving a gap in understanding how it can be applied to other popular models such as Vision Transformers (ViTs).
In this work, we investigate where and how to partially personalize a ViT model. Specifically, we empirically evaluate the sensitivity to data distribution of each type of layer. Based on the insights that the self-attention layer and the classification head are the most sensitive parts of a ViT, we propose a novel approach called FedPerfix, which leverages plugins to transfer information from the aggregated model to the local client as a personalization. Finally, we evaluate the proposed approach on CIFAR-100, OrganAMNIST, and Office-Home datasets and demonstrate its effectiveness in improving the model's performance compared to several advanced PFL methods. Code is available at \href{https://github.com/imguangyu/FedPerfix}{https://github.com/imguangyu/FedPerfix}
\end{abstract}

\section{Introduction} \label{sec:intro}
Federated learning (FL) \cite{mcmahan_communication-efficient_2017} has emerged as a promising method for training machine learning models on decentralized data without requiring direct data sharing. However, data heterogeneity among participating clients can present a significant challenge. Due to the various circumstances of the clients, the data across the clients can be non-independent and non-identically distributed (non-IID). 
Therefore, achieving satisfactory performance using a one-model-fits-all approach is difficult, and personalized models are often needed to achieve the best results. This has inspired the study of Personalized Federated Learning (PFL), where the focus is shifted from the performance of the global model on the server to the local models on the clients.


In the context of personalized federated learning, previous literature has explored two main approaches: \textit{full model personalization}~\cite{fallah_personalized_2020,deng_adaptive_2020,luo_pgfed_2022} and \textit{partial model personalization}~\cite{pillutla_federated_2022,oh_fedbabu_2022,li_fedbn_2021,collins_exploiting_2021,arivazhagan_federated_2019}. Full model personalization involves maintaining a separate local model for each client and updating it based on a joint objective, while partial model personalization aims to personalize only a subset of the model parameters. A convergence analysis~\cite{pillutla_federated_2022} suggests that partial model personalization can achieve most of the benefits of full model personalization with fewer shared parameters, offering advantages in terms of computation, communication, and privacy to enable the deployment of larger models on the clients.

However, recent literature shows that \textit{where} and \textit{how} to perform partial model personalization has a high correlation to the model architectures and the tasks~\cite{pillutla_federated_2022}, which requires further study when applied to a new architecture.
Despite the multitude of approaches proposed in the literature, the majority of methods have only been evaluated on Convolutional Neural Networks (CNNs). Meanwhile, Vision Transformers (ViTs)~\cite{dosovitskiy_image_2021} have demonstrated superior performance compared to CNNs in several tasks, such as image classification~\cite{pmlr-v139-touvron21a} and object detection~\cite{zhang_dino_2022,carion_end--end_2020}, making them an attractive option for personalized federated learning. However, to the best of our knowledge, the application of ViT in the federated learning community has received limited attention in the existing literature~\cite{qu_rethinking_2022}. 
Given the advantages of ViT shown under centralized training, it is reasonable to expect that these benefits can also be realized in PFL by offering a more robust model for improved performance on the clients.
Therefore, in this work, we investigate \textit{where} and \textit{how} to \textit{partially personalize} a \textit{ViT} model. 

Drawing from previous research on CNNs, layers that serve specific engineering purposes, such as feature extraction, normalization, or classification, have been identified as suitable candidates for partial model personalization~\cite{oh_fedbabu_2022,li_fedbn_2021,collins_exploiting_2021}. 
These layers might have a higher \textit{sensitivity} to the distribution of the training data. Therefore, aggregating the model weights trained from different data distributions may result in an inaccurate feature, while
keeping them updated locally can gain a better feature for the local data distribution.
Similarly, we select some candidates from ViT and conduct an empirical study to investigate the sensitivity of each type of layer. Specifically, we quantitatively evaluate the impact of keeping certain layers updated locally without aggregation with other clients. This evaluation shows that the \textit{self-attention layers} and the \textit{classification head} have a higher sensitivity than other layers, providing insights about \textit{where} to personalize.

To personalize the sensitive parts, 
one intuitive approach is to keep them completely local. 
However, the global aggregation has shown the capability to provide a more general global model than local models~\cite{mansour_three_2020}. Completely preventing the sharing with the global model will severely hinder the potential benefits of leveraging general knowledge from the aggregated global model. Therefore, we desire to train a personalization module to \ul{bridge the general knowledge and the client-specific knowledge.}

\begin{figure}[t]
\centering
\includegraphics[width=\linewidth]{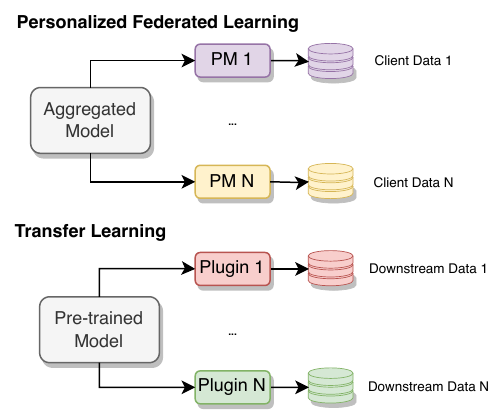}
\caption{\textbf{Analogy between transfer learning and personalized federated learning.} A pre-trained model can be transferred to different downstream data with different plugins. In personalized federated learning, we are seeking different personalization modules (PM) to transfer the aggregated model to different client data.}
\label{fig:intro}
\end{figure}

As existing works in transfer learning suggest, 
the same pre-trained model can be transferred to different downstream data by adding different tiny architectures~\cite{pfeiffer_adapterhub_2020,jia_visual_2022,li_prefix-tuning_2021,hu_lora_2021}. We refer to these tiny architectures as \textit{plugins} since they are small-sized parameters plugged into the pre-trained model. These plugins are trained while the weights of the pre-trained model are frozen. The plugged model can achieve comparable performance as the fully fine-tuned model~\cite{pfeiffer_adapterfusion_2021}. Since the model can maintain the same level of performance without changing the weights of the pre-trained model, the downstream-specific knowledge is captured by the plugins. Therefore, the plugins show the capability to \textit{personalize} the same model to satisfy different downstream data.
In federated learning, \ul{we can consider 
the aggregated model as an inferior version of the ``pre-trained'' model, and the data on different clients as the downstream data}, as shown in Fig.~\ref{fig:intro}.
With this perspective, we can therefore exploit the capability of the plugins in the federated learning context to capture client-specific knowledge for personalization.

Therefore, we select and adapt a specific family of plugin, \textit{Prefixes}, to personalize the self-attention layer and propose a novel approach \textbf{FedPerfix}, short for \textbf{Fed}erated \textbf{Per}sonalized Pre\textbf{fix}-tuning.

The main contributions of this paper are as follows:
\begin{itemize}
    \item 
    We perform an empirical study to reveal the sensitivity to data distribution of each type of layer in a ViT for PFL and locate the self-attention layer and classification head as the sensitive part to be personalized. (Section~\ref{sec:emp}) 
    \item We propose a novel partial model personalization approach on ViT, FedPerfix, inspired by the connection between PFL and transfer learning. Specifically, we exploit Prefix plugins to capture client-specific knowledge for personalization. (Section~\ref{sec:fedperfix})
    \item We conduct evaluations on CIFAR-100~\cite{cai_tinytl_2020}, OrganAMINIST~\cite{medmnistv2}, and Office-home~\cite{venkateswara2017deep}, which are across different domains, and degrees of data heterogeneity and achieve state-of-the-art performance compared with several competitive methods with lower resource requirements. (Section~\ref{sec:performance})
\end{itemize}



\section{Related Work}
\noindent \textbf{Personalized Federated Learning.} 
Personalized Federated Learning focuses on training a client-specific model to achieve satisfying performance on each client instead of a unified global model to fit all the client data. In terms of the shared parameters, full model personalization keeps a separate personalized model for each client and designs different training objectives, where the global model is usually served as the regularization term~\cite{tan_fedproto_2022,li_ditto_2021,zhang_personalized_2023,li_fedphp_2021,huang_personalized_2021, mendieta_local_2022}. The training objectives are designed with the idea of meta-learning~\cite{fallah_personalized_2020}, Moreau Envelopes~\cite{dinh_personalized_2022}, model interpolation~\cite{deng_adaptive_2020}, decoupling~\cite{chen_bridging_2022}, \etc. In contrast, partial model personalization focuses on personalizing some specific parameters inside a model. 
Some works simply keep these parameters, \eg, the classification head~\cite{arivazhagan_federated_2019, oh_fedbabu_2022}, and the batch normalization layers~\cite{li_fedbn_2021} updated locally, while some works design additional parameters, \eg, prototypes for each class~\cite{tan_fedproto_2022} or hypernetworks~\cite{shamsian_personalized_2021}, to further personalized these parameters. Our proposed method focuses on the self-attention layers in a ViT and leverages plugins in transfer learning, enabling personalization without introducing significant overhead. Our experimental results demonstrate that FedPerfix outperforms existing methods in terms of both efficiency and effectiveness.

\noindent \textbf{Vision Transformer.} 
ViTs, a type of attention-based neural network, have shown superior performance in several fields compared to CNNs. In particular, ViTs have achieved state-of-the-art performance on various benchmarks~\cite{dosovitskiy_image_2021,liu_video_2021,liu_swin_2021}. 
ViTs employ self-attention mechanisms to learn the dependencies between different regions of an image and can capture long-range dependencies more effectively than CNNs. 
Despite their success, the use of ViTs in federated learning has not been as widely explored as CNNs. Qu~\etal~\cite{qu_rethinking_2022} studied the performance of ViT under conventional FL settings and found that simply replacing CNNs with ViTs can greatly accelerate convergence and reach a better global model, especially when dealing with heterogeneous data, demonstrating the potential of ViTs in FL. 
In this paper, we extend the application of ViT to PFL and propose a novel approach to better personalize a ViT.

\noindent \textbf{Parameter-efficient Fine-tuning and Plugins.} 
Fine-tuning is a popular technique for adapting a pre-trained model to a new task. 
However, fine-tuning can be computationally expensive, especially for large models~\cite{he_towards_2022,yu_towards_2022}. Parameter-efficient fine-tuning (PEFT) has been proposed to address this issue by selectively updating only a subset of the model's parameters or a few added parameters. These added parameters are plugged into the pre-trained model; therefore, they are summarised as ``Plugins'' in this paper.
Although the location of the plugins for a transformer is various, including the input~\cite{jia_visual_2022}, the self-attention layer~\cite{li_prefix-tuning_2021}, and the feed-forward network~\cite{pfeiffer_adapterhub_2020}, all of them can be trained to achieve comparable performance as fine-tuning without updating the weights of the pre-trained model.
The plugins have also been applied to conventional FL to enable large pre-trained models and reduce the communication cost~\cite{sun_conquering_2022}. 
In the PFL, 
inspired by analogy from fine-tuning to the local training, the plugins can similarly be employed to personalize the representation from the global model to the local data distribution. Based on our empirical study, we choose prefixes~\cite{li_prefix-tuning_2021} since the parameters are directly applied to the self-attention layer and propose our method, FedPerfix.

\section{Method}
\subsection{Problem Formulation} 
We consider a classification problem in Computer Vision (CV).
A dataset $\mathcal{D}=\{(x,y)| x\in X, y\in Y \}$ is separated on $N$ clients, where $X$ is the input space and $Y$ is the label space. 
Data on each client is denoted as $\mathcal{D}_i$, and the distribution $P_i$. The client distributions are not identical. Each client has access to $m_i$ samples drawn IID from the distribution $P_i$. The total number of samples is $M=\sum_{i=1}^{N}m_i$. 
The hypothesis (model) on each client is noted as $h_i$, and the expected loss on the $i^{\text{th}}$ client is denoted as 
\begin{equation}
    \mathcal{L}_{\mathcal{D}_i}(h_i)=\mathbb{E}_{(x,y)\sim P_i}[\ell (h(x),y)]
\end{equation}
where $\ell$ is the loss function.
Further, considering the \textit{partial model personalization}, the parameters in each client model can be divided into two parts: global parameters $u$ and local parameters $v_i$, i.e.,  $h_i=u\cup v_i$. Specifically, $v_i = \varnothing$ indicates \textit{full model personalization}. 

In each communication round $t \in [T]$, each client will receive $u^{(t-1)}$ from the server and plug $v_i^{(t-1)}$ into $u^{(t-1)}$. Then, 
$u_i^{(t)}, v_i^{(t)}$ are trained on the local data.
After the local training, 
only the $u_i^{(t)}$ of $K$ sampled clients will be sent to the server and participate in the global aggregation. The sampled ratio (client participation rate) is denoted as $r=K/N$.
The global aggregation is formulated as 
\begin{equation}
         u^{(t)} = \sum_{i=1}^{K} \alpha_i u_i^{(t)}
\end{equation}
where $\alpha_i$ is the aggregation coefficient. Federated Averaging (FedAVG) is the most common aggregation algorithm where $\alpha_i = \frac{m_i}{M}$. After $T$ rounds of communication, each client model will receive a copy of the newest global parameters and plug its own local parameters into the global parameters. Then, the model will be evaluated on its own test data drawn IID from $P_i$.
To assess the overall performance, the mean and standard deviation of the \textbf{Top-1 classification accuracy (Acc)} for all clients will be reported.

\subsection{Recap: Partial Model Personalization for CNN} 
In this section, we provide a brief recap of the approaches for partial model personalization in CNNs. 
We aim to leverage the insights and experiences gained from previous research to explore the strategy to personalize ViTs partially. 

\textbf{FedRep~\cite{collins_exploiting_2021}}
is a partial model personalization method that aims to preserve client-specific information while leveraging common knowledge in the earlier layers of a CNN by keeping the last classification layer and a few blocks local and aggregating the rest.
\textbf{FedBN~\cite{li_fedbn_2021}}
is a partial model personalization method that updates Batch normalization (BN)~\cite{ioffe_batch_2015} layers locally while aggregating the rest of the model. This method leverages the local statistics of the BN layer to adapt to the data distribution of each client while maintaining a global representation of the model, achieving a better balance between personalization and global representation.
\textbf{FedBABU~\cite{oh_fedbabu_2022} } 
focuses on personalizing the classification head to address the issue of inconsistent feature spaces among clients. Unlike FedRep, FedBABU freezes the classification head and fine-tunes it for several steps before evaluation. By doing so, FedBABU can mitigate the negative impact of classification head drift and encourage consistent feature spaces across all clients.

In summary, previous approaches to partial model personalization for CNNs provide valuable insights to answer the ``where to personalize" question for ViTs: The layers to be personalized are usually designed with some engineering purpose, thus, are sensitive to data distribution. For instance, the normalization layers are designed to capture the statistical characters of the data, while classification heads are designed to map the feature to a predicted class. 
Building on these insights, we group the layers in ViT by their engineering purpose and conduct an empirical study to determine the layers in ViT that are the most sensitive to data distribution. We evaluated the performance of ViT while keeping some specific types of layers updated locally without aggregating with other clients. The results of our study are presented in Section~\ref{sec:emp}.

\begin{table}[t]

\caption{\textbf{Sensitivity to data distribution of each type of layer in a Vision Transformer.} 
The \textit{mean} and \textit{standard deviation} of the client's Top-1 Accuracy are reported for each type of layer (\textbf{Stand-alone}). Considering the classification head is the most sensitive type of layer, we also report the performance when a type of layer is kept updated locally along with the classification head (\textbf{Combined}). The \textbf{overall} performance is the mean of the stand-alone accuracy and combined accuracy.}
\label{tab:parts}
\begin{center}
\resizebox{\linewidth}{!}{%
\begin{tabular}{lccc}
\toprule
\centering
\bf Type &\bf Stand-alone &\bf Combined&\bf Overall
\\ \midrule
All Local& \res{34.74}{9.36} &-&-\\
All Global& \res{23.29}{11.31} &-&-\\
Classification Head& \res{44.42}{7.98}&-&-\\
\midrule
Patch Embedding&  \res{27.61}{10.02}&\res{43.45}{8.69}& $35.53$\\
Position Embedding & \res{23.80}{11.42}& \res{45.04}{8.08} & $34.42$\\
LayerNorm&\res{23.94}{11.19}& \res{44.60}{7.96}& $34.27$\\
Self-attention& \res{42.95}{8.68}& \res{44.63}{8.67}& $\mathbf{43.79}$\\
MLP&\res{42.53}{8.82}&\res{42.63}{8.84}& $42.58$\\

\bottomrule
\end{tabular}%
}
\end{center}
\end{table}

\begin{figure*}[t]
\centering
\includegraphics[width=\linewidth]{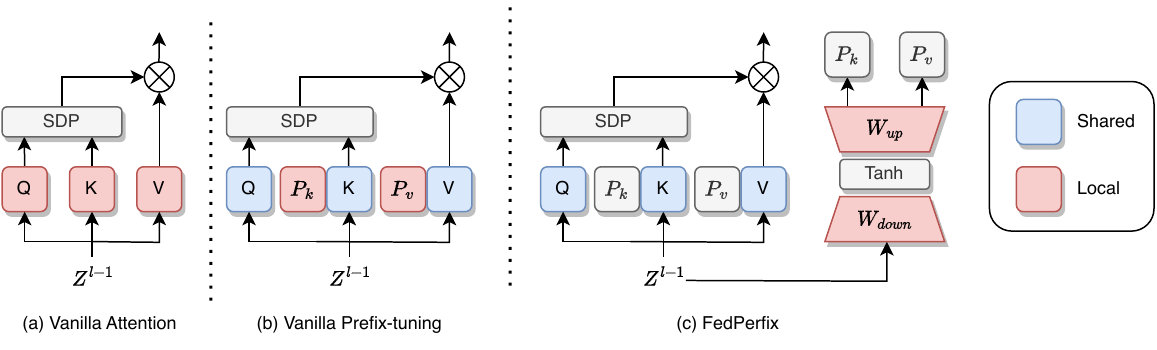}
\caption{\textbf{Several designs to personalize the self-attention layer in a ViT.} SDP is short for Scaled Dot Product, the red part indicates the local parameters, the blue part indicates the global parameters and the gray parts are vectors or modules with no learnable parameter. (a) Vanilla attention simply keeps the self-attention and classification updated locally. (b) Vanilla prefix-tuning only keeps the prefixes updated locally and aggregates the original self-attention layer. (c) FedPerfix uses a local adapter to generate the prefixes and aggregates the original self-attention layer. The adapter is composed of a scale-down, an activation, and a scale-up layer.}
\label{fig:prefix}
\end{figure*}

\subsection{Empirical Study: Partial ViT Personalization}\label{sec:emp}
In a ViT, we consider the following layers as candidates
to be personalized and evaluate their sensitivity to data distribution when they are updated locally:
\textbf{Patch Embedding} layer converts the input image to a sequence of patches, then applies a linear projection to map each patch into an embedding vector.
\textbf{Position Embedding} layer is responsible for injecting positional information into the sequence of patches produced by the patch embedding layer, allowing the model to capture the spatial arrangement of the image.
\textbf{LayerNorm} is a normalization layer commonly used in ViT to normalize the feature maps across the channel dimension.
\textbf{Self-attention layer} allows the model to attend to different regions of the input image and capture long-range dependencies between them.
\textbf{MLP} layers, following the self-attention layer, enable the model to capture non-linear relationships between the image patches.
\textbf{Classification Head} applies a linear projection from the [CLS] token to a predicted class.

We perform experiments on CIFAR-100 with our default setting, which will be introduced in Section~\ref{sec:setting}. We keep the specified type of layer updated locally and report the mean and standard deviation of the Top-1 Accuracy across all clients as the metrics to indicate the \textit{sensitivity} to data distribution.
The result is shown in the second column of Table~\ref{tab:parts}. 

From the table, we can draw several insights. Keeping all layers local yields better performance than aggregating all layers. This suggests that aggregation of heterogeneous data may have a negative impact on some layers. Meanwhile, keeping each type of layer updated locally allows for greater adaptability to each client's data, leading to different degrees of improvement compared with keeping every layer global. 
Among all types of layers, the classification head is the most sensitive to data distribution, leading to the highest stand-alone performance, which is as expected.
Except for the classification head,
self-attention layer is more sensitive to data distribution than other components, showing a higher standard-alone performance.

Considering the vital importance of the classification head, we also reported the combined performance when a type of layer is kept updated locally along with the classification head,
as shown in the third column of Table~\ref{tab:parts}. To decide the most sensitive layers to be personalized, we averaged the performance under the two settings as the overall performance.

Based on the overall performance, personalizing the self-attention and classification head is one possible answer to the question of \textit{where} to personalize. Besides, a vanilla baseline to keep the self-attention layers and classification head updated locally is proposed, which is referred to as Vanilla Attention as described in Fig.~\ref{fig:prefix} (a). 


\subsection{Proposed Baseline: Vanilla Prefix-tuning}\label{sec:method}

Building upon the insights about \textit{where} to personalize gained in the previous section, we move to the next question about \textit{how} to personalize the selected layers. As obtained from our empirical study, Vanilla Attention is a baseline to personalize the self-attention layer by completely keeping the self-attention updated locally. However, it will also prevent it from learning general information for the global model. 

As explained in Section \ref{sec:intro}, the plugins can be trained on the local data to capture the client-specific knowledge as a personalization module. There are three widely used plugins for ViTs: a) \textbf{Prompts} that are learnable embeddings appended to the inputs, b) \textbf{Adapters} that are inserted into the MLP layers, and c) \textbf{Prefixes} that are appended to the key and value matrix of the self-attention layer. Given that the self-attention layer is one of the most sensitive layers, 
It is plausible that the Prefixes integrated nearest to the self-attention layer are capable of capturing the most relevant client-specific information pertaining to the self-attention mechanism.
Besides, we further compare the effectiveness of all three plugins and discuss it in Section~\ref{sec:design}. Therefore, we select Prefixes as our personalization module. 

Specifically, learnable prefixes are appended to the self-attention layer and kept updated locally on the client. Meanwhile, the original self-attention layer is shared across all clients as normal to capture global dependencies. We refer to this approach as Vanilla Prefix-tuning.

An illustration of how Prefixes cooperate with the self-attention layer is shown in Fig.~\ref{fig:prefix} (b). Besides, we formulate the output of one head of the self-attention layer as
\begin{equation}
\label{eq:prefix}
\begin{aligned}
    head_i = &Attention(\mZ^{l-1}\mW_q^{(i)}, 
    \\&\mZ^{l-1}[\mP_k,\mW_k^{(i)}], 
    \\&\mZ^{l-1}[\mP_v,\mW_v^{(i)})])
\end{aligned}
\end{equation}
where $\mZ^{l-1}$ is the output from the last layer, $\mP_k$ and $\mP_v$ are prefixes, $\mW_q^{(i)}, \mW_k^{(i)},$ and $ \mW_v^{(i)}$ are the parameters in the self-attention layer, and $[,]$ is the concatenate operation.

\subsection{FedPerfix: Stabilize the Prefix}\label{sec:fedperfix}
In Vanilla Prefix-tuning, the prefixes are randomly initialized, which can result in unstable performance when initialized with different weights. We later conduct an experiment to demonstrate such instability in Section~\ref{sec:design} when Vanilla Prefix-tuning is initialized with different weights.
To address this issue, we draw inspiration from the parallel attention design~\cite{yu_towards_2022}, which uses adapters to stabilize the prefixes. Similarly, we propose to use adapters in Vanilla Prefix-tuning to stabilize the prefix initialization. Specifically, we add a parallel adapter to prepare the prefixes for each layer, as shown in Fig.~\ref{fig:prefix}. Meanwhile, we add a hyperparameter $s$ to control the efficiency of the adapter. Following the design of parallel attention, the prefixes are generated as
\begin{equation}
\small
\mP_k, \mP_v = Adapter(\mZ^{l-1})=\text{Tanh}(\mZ^{l-1}\mW_{down})\mW_{up}
\end{equation}
where Tanh is the activation function, $\mW_{down}$ and $\mW_{up}$ are parameters of the scaling layers of the adapter.
As a result, the output of one head of the self-attention layer with parallel attention is shown in Fig.~\ref{fig:prefix}(c) and formulated as 
\begin{equation}
\begin{aligned}
    head_i = &Attention(\mZ^{l-1}\mW_q^{(i)}, 
    \\&\mZ^{l-1}[s\mP_k,\mW_k^{(i)}], 
    \\&\mZ^{l-1}[s\mP_v,\mW_v^{(i)}]).
\end{aligned}
\end{equation}

 \begin{table*}[t]
\caption{\textbf{Performance and required resources for each method.} The mean and standard deviation of the Top-1 Accuracy among all clients are reported. The bold style indicates the best performance in each dataset.}
\label{tab:main}
\begin{center}
\resizebox{\linewidth}{!}{
\begin{tabular}{c|ccc|ccc}
\toprule
\centering
\multirow{2}{*}{\bf Method}&\multicolumn{3}{c|}{\bf Performance}& \bf Storage & \bf Computation & \bf Communication
\\&CIFAR-100&OrganAMNIST&Office-Home&(\# Params)&(FLOPs)&(\# Params)\\
\midrule
FedAvg &\res{23.29}{11.31}&\res{87.31}{5.98}&\res{21.47}{6.24} &21.03M (100\%)&65.65M (100\%)&21.03M (100\%)\\
Local  &\res{34.74}{9.36}&\res{78.26}{11.07}&\res{20.39}{7.26}&21.03M (100\%)&65.65M (100\%)&0 (0\%)\\
\midrule
APFL&\res{44.88}{10.50}&\res{89.74}{5.83}&\res{24.23}{7.02}&42.06M (200\%)&131.30M (200\%)&21.03M (100\%)\\
Per-FedAVG &\res{33.86}{8.01}&\res{82.81}{7.13}&\res{17.09}{4.83}&21.03M (100\%)&131.30M (200\%)&21.03M (100\%)\\
\midrule 
FedBN& \res{23.94}{11.19}&\res{87.63}{5.78}&\res{21.25}{5.89}&21.03M (100\%)&65.65M (100\%)&21.01M (100\%)\\
FedBABU& \res{41.41}{8.87}&\res{88.38}{7.16}&\res{19.50}{7.71}&21.03M (100\%)&65.65M (100\%)&20.66M (98\%)\\
FedRep& \res{44.42}{7.80}&\res{92.63}{3.77}&\res{23.67}{5.97}&21.03M (100\%)&65.65M (100\%)&20.66M (98\%)\\
\midrule
Vanilla Attention& \res{44.63}{8.67}&\res{88.90}{5.87}&\res{22.55}{6.37}&21.03M (100\%)&65.65M (100\%)&13.89M (66\%)\\
FedPerfix (ours)&\res{\mathbf{48.10}}{7.76} &\res{\mathbf{93.17}}{3.51}&\res{\mathbf{24.38}}{8.47}&24.42M (116\%)&66.58M (101\%)&20.66M (98\%)\\
\bottomrule
\end{tabular}
}
\end{center}
\end{table*}

In conclusion, FedPerfix is proposed as a novel approach to transfer the information from the aggregated self-attention layer to fit the local data better, leveraging the plugins in transfer learning. 
The local prefixes are trained for personalization, while the global self-attention layer captures global dependencies. Therefore, 
FedPerfix provides a promising solution to address the question of how to perform personalization for ViTs.

\section{Experiment}
\subsection{Experiment Details}\label{sec:setting}
\noindent\textbf{Dataset.} We use three popular datasets to evaluate the performance: \textbf{CIFAR-100}~\cite{krizhevsky_learning_2012}, \textbf{OrganAMNIST}~\cite{medmnistv2}, and \textbf{Office-Home}~\cite{venkateswara2017deep}. CIFAR-100 is a widely-used image classification dataset consisting of 50,000 RGB images across 100 classes. Office-Home is a domain adaptation dataset for object recognition tasks, which contains 15,500 images across 65 categories, captured from four different domains: Artistic images, Clipart, Product images, and Real-World images. OrganAMNIST is a medical image classification dataset that includes 58,850 gray-scale images of organs and tissues from human anatomy. The images are classified into 11 different classes. These datasets are chosen to evaluate the performance in various scenarios, including different data scales, domains, and the number of classes.

\noindent\textbf{Federated learning settings.} 
We conduct $T=50$ communication rounds with each local training consisting of 10 epochs. To account for the various factors that can affect federated learning performance, we adjust the default settings for each dataset to focus on different scenarios with varying degrees of data heterogeneity of \textbf{label skew} and \textbf{concept skew}. For CIFAR-100, which has a relatively high number of samples and classes, we simulate a federated learning environment with $N=64$ clients using a Dirichlet distribution with $\alpha=0.1$ and sampling $K=8$ clients with a ratio $r=0.125$ for each round.
For OrganAMNIST, which has a similar number of samples but fewer classes than CIFAR-100, we partition the data using a Dirichlet distribution with a larger $\alpha=0.5$ and apply the same client settings as CIFAR-100.
As for Office-Home, which has fewer samples captured from 4 different domains, we focus on concept skew by partitioning the data from each domain into four different clients using a Dirichlet distribution with $\alpha=1.0$, resulting in a total of $N=16$ clients. In each communication round, we randomly sample $K=4$ clients from any domain with a ratio $r=0.25$. A visualization of the data partitioning is provided in \textcolor{blue}{Supplementary A.1}.

\noindent\textbf{Model Architecture and Baselines.} We choose ViT-Small (ViT-S)~\cite{dosovitskiy_image_2021} with a patch size of $16$ and an image size of $224$ as our model for evaluation. We compare our proposed FedPerfix method against a range of baseline methods, including widely used approaches like FedAVG and local training without aggregation (Local). We also compare our method against advanced full model personalization methods, namely APFL~\cite{deng_adaptive_2020} and Per-FedAVG~\cite{fallah_personalized_2020}, and three partial model personalization methods, namely FedBN, FedRep, and FedBABU. Besides, we also include the performance of Vanilla Attention, which keeps the self-attention layer and the classification head updated locally, as a reference. APFL adapts the global and local model through an adaptive mixture coefficient, while Per-FedAVG employs meta-learning to improve the global model. These methods are model-agnostic and can be applied to ViTs. FedBN keeps the batch normalization layer updated locally and aggregates the remaining layers in the server with FedAVG. To adapt it to ViTs, we replace the batch normalization layer with the layer normalization layer. FedRep keeps the classification head updated locally while aggregating the other layers in the server. FedBABU freezes the classification head and aggregates the remaining layers in the server, then fine-tunes the classification head on the local data for one step before the evaluation. 

\noindent\textbf{Evaluation Metrics.} We report the final mean and standard deviation of Top-1 accuracy among all clients after all communication rounds finish as the evaluation metrics. Besides, to evaluate the feasibility of the methods, an analysis of storage, computation, and communication costs is reported in Section~\ref{sec:cost}.

\noindent\textbf{Implement Details.} All hyperparameters in each method are tuned as optimal in a range. Each model is optimized with the SGD~\cite{ruder_overview_2017} optimizer with the optimal learning rate, which is $0.01$ for most methods. The server and all the clients are simulated in one machine, and we resize the images in all datasets as $224\times224$ RGB images to fit the model and set the batch size of the client data as $64$. The model is implemented from the TIMM~\cite{wightman_pytorch_2019} library.
All the experiments are implemented in Pytorch~\cite{NEURIPS2019_9015} and performed on 4 Nvidia A5000 GPUs. More details are provided in \textcolor{blue}{Supplementary A.2}.

\subsection{Performance Evaluation}\label{sec:performance}
Based on the evaluation results presented in Table~\ref{tab:main}, several insights can be drawn regarding the performance of the compared methods. It can be observed that FedAVG performs worse than Local when facing a large label skew, \ie, on CIFAR-100. This indicates the importance of addressing data heterogeneity in federated learning scenarios. However, when the label skew is not that extreme, the information gathered from the global aggregation can alleviate the overfitting due to few training samples on the client, leading to higher performance.

In addition, it is worth noting that the methods that focus on the global objective instead of information transferring from the global to the client model, such as Per-FedAVG, achieve better performance than FedAVG but still show inferior performance compared to Local when facing extreme label skew or concept skew, \ie, CIFAR-100 and Office-Home. This suggests that direct modification of the global objective may not always be the most effective way when using ViTs. Performing meta-learning on relatively small-size client data for a relatively large ViT is a challenging task, which limits the effectiveness of Per-FedAVG. 

On the other hand, how to leverage the information provided in the global model is crucial for partial model personalization.
The layer normalization version of FedBN shows little impact on the performance, leading to a negligible improvement based on FedAVG. However, APFL, FedRep, Vanilla Attention, FedBABU, and FedPerfix find a more suitable way to transfer the information from the global model, leading to superior performance than other methods. Consistent with the conclusion we draw from the empirical study, FedRep, FedBABU, and Vanilla Attention personalized the sensitive parts in ViT, thus outperforming FedAVG and Local by a significant margin. 
As an extension, APFL and our method, FedPerfix, balance the information between the local and global models in a proper manner, leading to the highest two performances considering the performance across all three datasets.

Surprisingly, we find that there is a connection between FedPerfix and APFL. Equation~\ref{eq:prefix}, which shows the output of a self-attention with Prefixes, can be rewritten as 
\begin{equation}
\begin{aligned}
    head(x) = 
    &(1-\lambda(x))\underbrace{Attn(x\mW_q, 
x\mW_k, 
x\mW_v)}_{\text{aggregated attention}}\\
&+ \lambda(x)\underbrace{Attn(x\mW_q, 
x\mP_k, 
x\mP_v)}_{\text{personalized attention}}
\end{aligned}
\end{equation}
where $\lambda(x)$ is a scalar representing the normalized attention weights on the prefixes~\cite{he_towards_2022}. More details are provided in \textcolor{blue}{Supplementary B}. It reveals that \textit{the local prefixes are learned as a mixture coefficient between the aggregated attention from the global model and the personalized attention from the prefixes}. The high performance of both APFL and FedPerfix indicates the effectiveness of a mixture between the local and global models in PFL. Compared with APFL, FedPerfix only mixed the sensitive self-attention layers instead of the entire model, leading to higher performance. Meanwhile, FedPerfix trains the Prefix, which can be considered as the mixture coefficient, and the model simultaneously, while APFL needs separate training for the personalized model. Therefore, FedPerfix can achieve higher performance with fewer \textbf{storage}, \textbf{computation}, and \textbf{communication} resources. Furthermore, we provide a more detailed resource requirements analysis of each method in the following Section~\ref{sec:cost}.
\subsection{Client-wise Performance}
\begin{figure}[t]
\centering
\includegraphics[width=\linewidth]{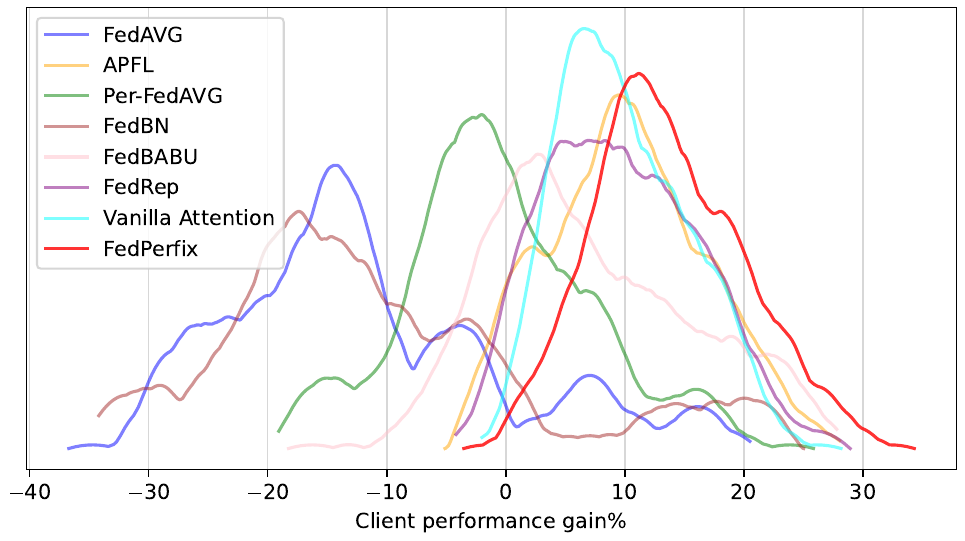}
\caption{\textbf{Client-wise performance for each method on CIFAR-100.} Density of the accuracy gain compared with Local training is plotted. FedPerfix provides highest upper and lower bounds among all the methods.}
\label{fig:client}
\end{figure}

In Fig.~\ref{fig:client}, we present a density plot depicting the performance gain of each method in comparison to Local. Notably, FedPerfix demonstrates an average performance gain of over 10\%, surpassing all other methods. Moreover, FedPerfix outperforms all other methods by achieving up to 30\% performance gains for certain clients. Conversely, some clients experience performance degradation under other methods, while FedPerfix ensures performance gains for nearly all clients. 
In conclusion, FedPerfix shows a \textit{higher upper bound} while maintaining \textit{a high lower bound}, which will encourage more clients to involve in the federated training with performance-gaining guarantees for almost all clients.

\subsection{Resource Requirements Analysis}\label{sec:cost}

In federated learning, it is crucial to consider the practicality of each method in terms of the storage, computation, and communication resources required by each participating client. As such, it is imperative to analyze the resource demands of each approach to determine its feasibility in realistic federated settings.

We present a resource analysis of the evaluated methods, taking into account the storage, computation, and communication demands, which are crucial aspects in practical federated learning scenarios where clients may have limited resources. Table~\ref{tab:main} reports the parameter size required to store the model, the FLOPs needed for training, and the parameter size required for communication. Notably, FedPerfix achieves superior performance compared to other methods with slightly more resources for storage and computation but fewer resources for communication. 

Overall, our proposed FedPerfix approach demonstrates the highest efficiency in terms of achieving superior performance while utilizing fewer storage, computation, and communication resources compared to other methods. These findings highlight the potential of our approach to effectively address the resource constraints often encountered in realistic federated learning scenarios.


\section{Ablation Study}

In this section, we conduct an ablation study on the CIFAR-100 dataset to evaluate the robustness of FedPerfix under various federated learning settings and to investigate FedPerfix with different designs.

\begin{table}[t]
\caption{{Performance on CIFAR-100 with different backbones.}}
\label{tab:cnn}
\begin{center}
\begin{tabular}{c|cc}
\toprule
\centering
\bf Method&\bf ResNet50 &\bf ViT-Small\\
\midrule
FedAVG&\res{17.87}{13.16}&\res{23.29}{11.31}\\
Local&\res{28.34}{9.83}&\res{34.74}{9.36}\\
APFL&\res{30.71}{9.89}&\res{44.88}{10.50}\\
Per-FedAVG&\res{28.82}{8.53}&\res{33.86}{8.01}\\
FedBN&\res{19.33}{10.25}&\res{23.94}{11.19}\\
FedBABU&\res{24.56}{7.82}&\res{41.41}{8.87}\\
FedRep&\res{39.52}{10.92}&\res{44.42}{7.80}\\
\textbf{FedPerfix}& - & \res{48.10}{7.76}\\
\bottomrule
\end{tabular}%
\end{center}
\end{table}

\subsection{Performance Comparison with CNN Backbone}
To further demonstrate the motivation behind our work, \ie, addressing PFL with ViT, we conduct experiments to compare ViT and CNN as different backbones for the same set of methods. Given that the comparison methods are \textit{originally designed with a CNN backbone}, we conduct experiments on CIFAR-100 with a CNN backbone as a reference to investigate the impact of replacing the CNNs with ViTs in these methods. To fairly compare the performance, we choose ResNet50~\cite{he_deep_2016} as the CNN backbone due to its similar parameter size (\textbf{24.37M}) to ViT-Small~\cite{dosovitskiy_image_2021} (\textbf{21.03M}). 

The result is shown in Table~\ref{tab:cnn}. Simply replacing the CNN backbone with the ViT backbone  significantly improves performance for each method without additional operations. It suggests that the advantages of ViTs that demonstrated superior performance compared to CNNs in centralized settings may also translate to personalized federated learning scenarios. 

The success of ViTs in existing methods highlights their potential for personalized federated learning. However, there is a lack of approaches specifically designed for ViTs. To address this gap, we propose FedPerfix, which is tailored for ViTs and focuses on personalizing the sensitive parts of the network. By doing so, FedPerfix further improves performance compared to existing methods that incorporate ViTs. Our approach fills an important research gap and suggests that there is still much to be gained from exploring the unique properties of ViTs for personalized federated learning.

\subsection{Different FL Settings}
In order to assess the robustness of FedPerfix in more challenging federated learning scenarios, we conduct experiments that involve varying the number of clients with fewer samples per client and a lower client participation rate. We also compare the performance of FedPerfix with other competitive methods and report the results in Table~\ref{tab:fl}. Our experimental results demonstrate that FedPerfix exhibits robust performance in different federated learning settings, outperforming all other methods consistently. Thus, FedPerfix can be considered a reliable solution for extreme federated learning scenarios.

 \begin{table}[t]
\caption{Performance on CIFAR-100 under different federated learning settings.}

\label{tab:fl}
\begin{center}

\resizebox{\linewidth}{!}{
\begin{tabular}{c|cc|cc}
\toprule
\centering
\multirow{2}{*}{\bf Method}&\multicolumn{2}{c|}{$\mathbf{N=64}$}& \multicolumn{2}{c}{$\mathbf{N=128}$}
\\&$r=6.25\%$&$r=12.5\%$&$r=6.25\%$&$r=12.5\%$\\
\midrule
FedAvg &\res{19.76}{9.35}&\res{23.29}{11.31}&\res{19.64}{10.01}&\res{23.29}{11.81}\\
Local&\res{32.35}{11.64}&\res{34.74}{9.36}&\res{31.79}{12.00}&\res{36.03}{9.57}\\\
APFL&\res{42.55}{13.51}&\res{44.88}{10.50}&\res{41.64}{12.77}&\res{43.92}{9.40}\\
FedRep&\res{35.72}{10.33}&\res{44.42}{7.80}&\res{35.88}{9.59}&\res{44.70}{8.40}\\
\textbf{FedPerfix}&\res{\mathbf{43.80}}{11.56}&\res{\mathbf{48.10}}{7.76}&\res{\mathbf{43.96}}{10.42}&\res{\mathbf{46.96}}{9.00}\\
\bottomrule
\end{tabular}
}
\end{center}
\end{table}

\subsection{Investigation of FedPerfix}\label{sec:design}
In this subsection, we investigate the impact of the model size of FedPerfix and verify its effectiveness compared with other designs. 

\noindent\textbf{Impact of Model Size.} The model size is a crucial factor in federated learning, as it determines the resource requirements for storage, computation, and communication.
To examine the impact of model size on the performance of FedPerfix, we conducted experiments with three different ViT backbones: ViT-Tiny, ViT-Small (Default), and ViT-Base. Additionally, we included the results of the baselines and two competitive methods as a reference. As shown in Table~\ref{tab:size}, FedPerfix still yields superior performance with different model sizes. Besides,
the larger models generally performed better, with ViT-Base achieving the highest performance for each method. However, we note that the performance improvement decreases as the model size increases, suggesting a potential trade-off between model size and performance. This finding highlights the importance of selecting an appropriate model size to achieve the optimal balance between performance and resource requirements, particularly in resource-constrained environments where storage and computation resources are limited. 


\noindent\textbf{FedPerfix vs. Vanilla Prefix-tuning.} As mentioned in Section~\ref{sec:fedperfix}, the initialization of the Prefixes will affect the performance. To show the effectiveness of the parallel attention design in FedPerfix, we conduct experiments with Vanilla Prefix-tuning under two different initialization, initialized with zero (Prefix-Z) and random initialization (Prefix-R). Then, we compare the results with FedPerfix. The result is shown in Table~\ref{tab:design}.
 As expected, the prefixes generated from the parallel attention yield better performance than the other two manually initialized prefixes,
 highlighting the effectiveness of the parallel attention design in FedPerfix.

\noindent\textbf{FedPerfix vs. Prompts \& Adapters. } Prompts and adapters are also popular plugins that are widely used. To verify whether FedPerfix is the most effective among all families of the plugins, we also apply Prompts and Adapters to personalized federated learning. Specifically, we append several trainable Prompts to the input embeddings and keep them updated locally as the implementation of Prompt-tuning~\cite{jia_visual_2022}. Meanwhile, we add Adapters to the MLP layers as the implementation of Adatper-tuning~\cite{pfeiffer_adapterhub_2020}. The results are shown in Table~\ref{tab:design}. Adding personalized Prompts doesn't improve the performance compared with simply keeping the classification head updated locally, indicating that the transformation to the input is not effective in PFL. However, adding Adapters to the MLP layers can also lead to a promising result as FedPerfix. Recap the result of our empirical study shown in Table~\ref{tab:parts}, MLP layers are also sensitive to data distribution in a ViT. We note that as another instance to show the effectiveness of adding plugins to the sensitive parts of a ViT in PFL.  
\begin{table}[t]
\caption{{Performance on CIFAR-100 with different model sizes.}}
\label{tab:size}
\begin{center}
\begin{tabular}{c|ccc}
\toprule
\centering
\bf Method&\bf ViT-Tiny &\bf ViT-Small &\bf ViT-Base\\
\midrule
FedAVG&\res{19.56}{10.27}&\res{23.29}{11.31}&\res{24.82}{11.53}\\
Local&\res{33.73}{9.20}&\res{34.74}{9.36}&\res{38.93}{10.08}\\
APFL&\res{37.66}{10.04}&\res{44.88}{10.50}&\res{47.57}{9.57}\\
FedRep&\res{41.58}{7.19}&\res{44.42}{7.80}&\res{44.34}{8.35}\\
\bf FedPerfix&\bf $\mathbf{44.71}_{\pm 8.47}$&\res{\mathbf{48.10}}{7.76}& $\mathbf{48.40}_{\pm 8.18}$\\
\bottomrule

\end{tabular}%
\vspace{-8mm}
\end{center}
\end{table}

\begin{table}[h]
\caption{{Performance of different designs on CIFAR-100.} Prefix-Z and Prefix-R mean vanilla prefix-tuning with zero and random initialization. }
\label{tab:design}
\begin{center}
\resizebox{\linewidth}{!}{%
\begin{tabular}{c|cc|cc}
\toprule
\centering
\bf FedPerfix&\bf Prefix-Z &\bf Prefix-R &\bf Prompts &\bf Adapters\\
\midrule
$\mathbf{48.10}_{\pm 7.76}$&$47.37_{\pm 8.47}$&$46.98_{\pm 8.10}$&$44.19_{\pm 8.13}$&$47.99_{\pm 8.59}$\\
\bottomrule
\end{tabular}%
}
\end{center}
\end{table}



\section{Conclusion}

In this work, we studied two research questions of where and how to personalize a ViT in federated learning. We conducted an empirical study to reveal that the self-attention and classification layers are the most sensitive layers for personalization. Based on that, we proposed FedPerfix, a novel method that introduces Prefixes with parallel attention to personalize the self-attention layers. Through extensive experiments on various datasets with different degrees of data heterogeneity, we demonstrated that FedPerfix achieves state-of-the-art performance while also reducing resource requirements. Our work focuses on ViTs and represents a shift in attention from the extensively researched CNNs, serving as inspiration for further investigation.

\section{Acknowledgement}
This work is partially supported by the NSF/Intel Partnership on MLWiNS under Grant No. 2003198.

\clearpage
{\small
\bibliographystyle{ieee_fullname}
\bibliography{references}
}
\clearpage

\begin{figure*}[t]
     \centering
     \begin{minipage}[b]{0.33\linewidth}
         \centering
         \includegraphics[width=\linewidth]{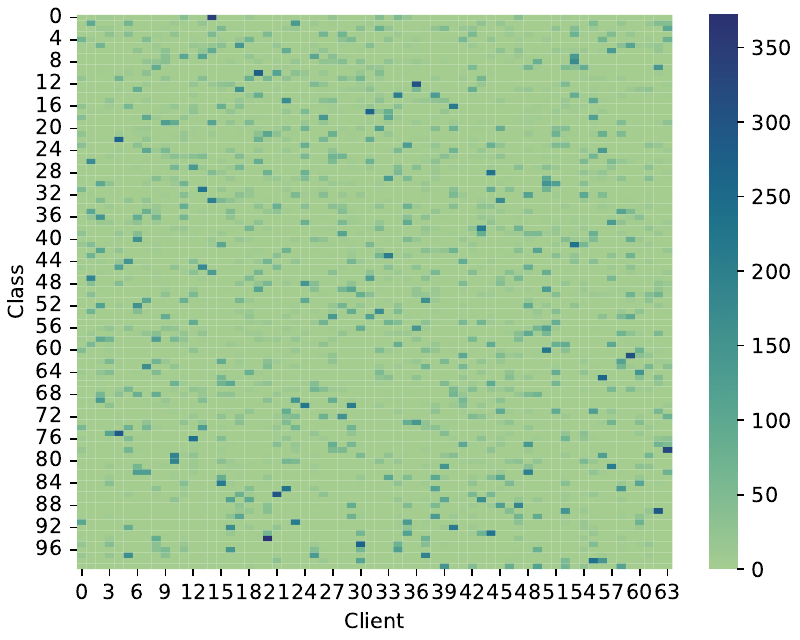}
         {\footnotesize(a) CIFAR-100}
     \end{minipage}
     \hfill
     \begin{minipage}[b]{0.33\linewidth}
         \centering
         \includegraphics[width=\linewidth]{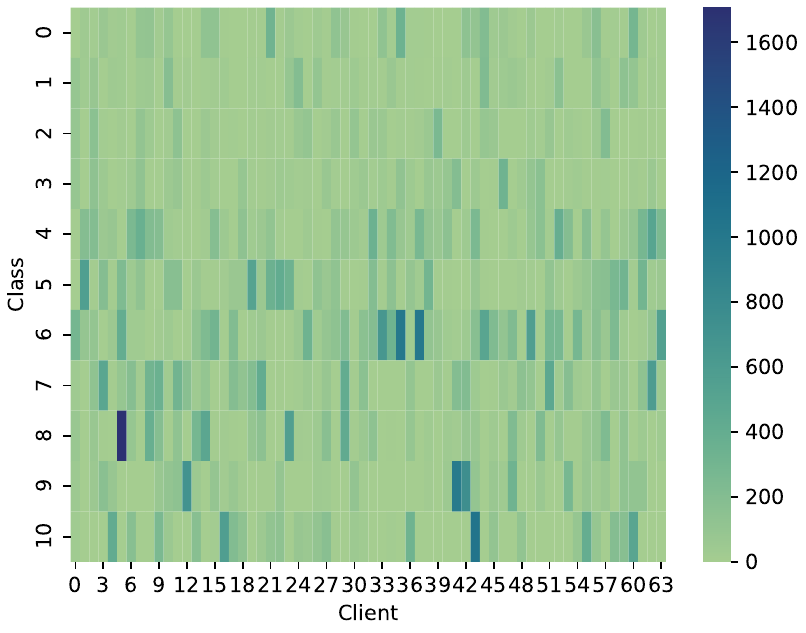}
         {{\footnotesize(b) OrganAMNIST}}
     \end{minipage}
     \hfill
     \begin{minipage}[b]{0.33\linewidth}
         \centering
         \includegraphics[width=\linewidth]{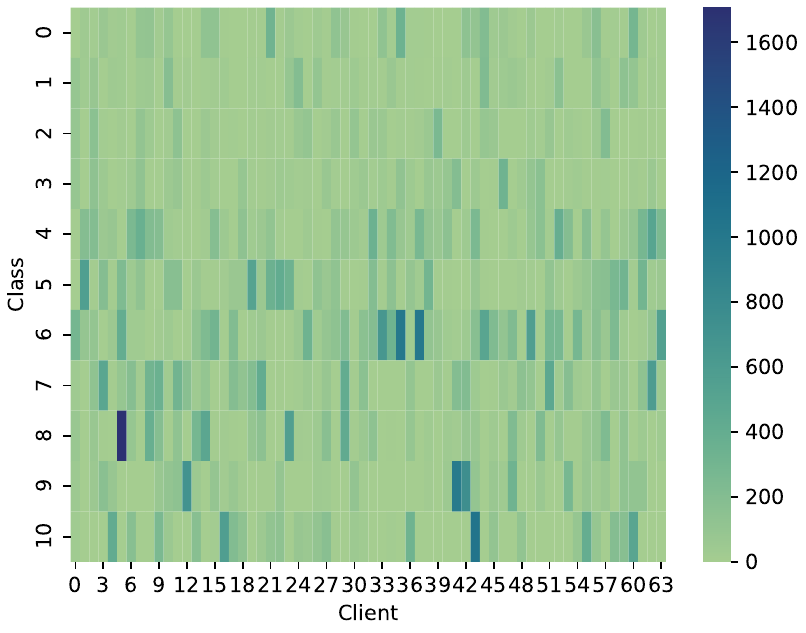}
         {{\footnotesize(c) Office-Home}}
     \end{minipage}
\caption{Data Partitioning of each dataset.}
\label{fig:data}
\end{figure*}
\begin{figure*}[h]
     \centering
     \begin{minipage}[b]{0.33\linewidth}
         \centering
         \includegraphics[width=\linewidth]{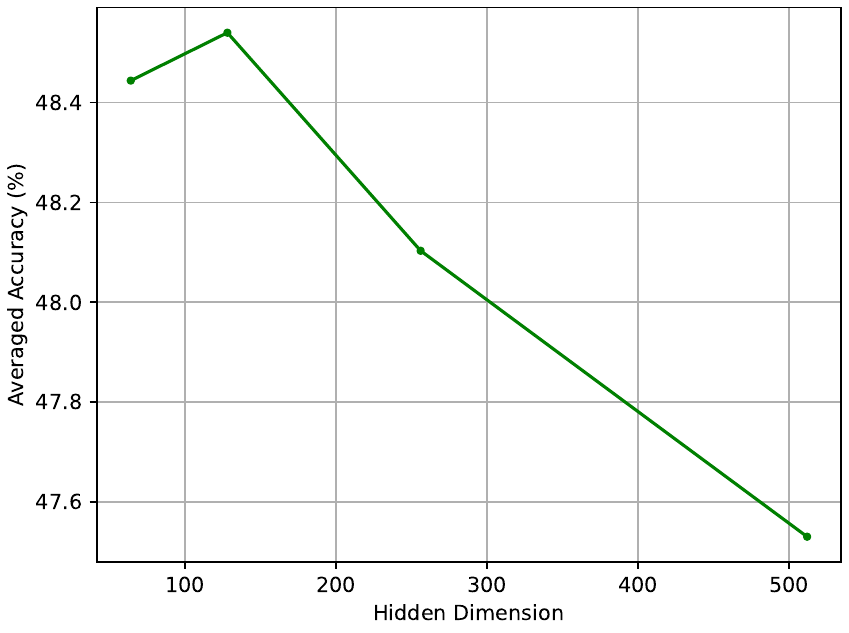}
         {\centering \footnotesize (a) Hidden Dimension vs. Performance}
     \end{minipage}
     \hfill
     \begin{minipage}[b]{0.33\linewidth}
         \centering
         \includegraphics[width=\linewidth]{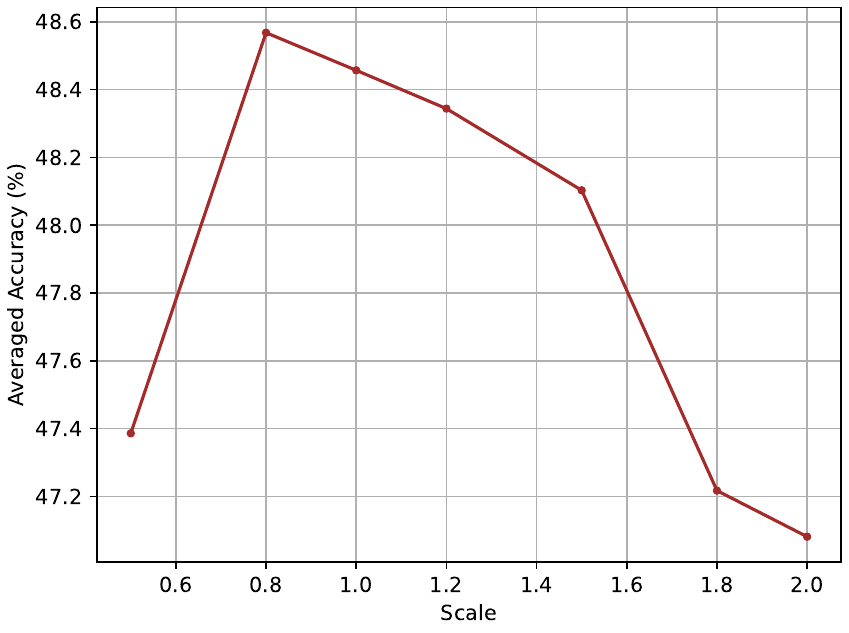}
         {\centering{\footnotesize(b) Scale vs. Performance}}
     \end{minipage}
     \hfill
     \begin{minipage}[b]{0.33\linewidth}
         \centering
         \includegraphics[width=\linewidth]{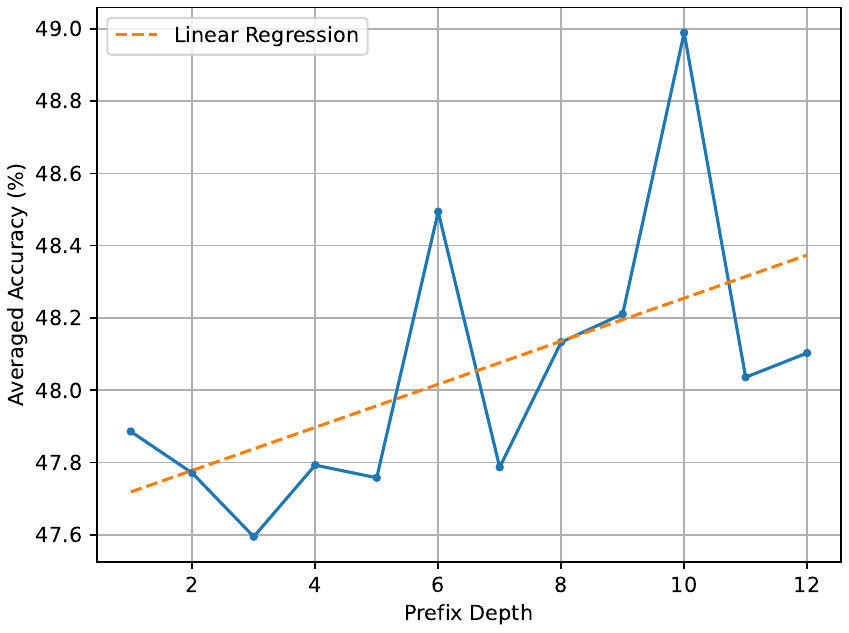}
         {{\footnotesize(c) Prefix Depth vs. Performance}}
     \end{minipage}
\caption{Impact of each hyperparameter on CIFAR-100.}
\label{fig:ablation}
\end{figure*}
\appendix
\section*{Supplementary}
The supplementary is organized in the following sections:
\begin{itemize}
    \item Section~\ref{sec:exp}: More details about the experiments.
    \item Section~\ref{sec:apfl}: Connection and comparison between APFL~\cite{deng_adaptive_2020} and FedPerfix.
    \item Section~\ref{sec:ab}: More hyperparameter ablations of FedPerfix.
\end{itemize}

\section{Experiment Details}\label{sec:exp}
\subsection{Visualization of Data Partitioning}
We partition the data in each dataset with Dirichlet distributions, which is a common setting in previous work~\cite{mendieta_local_2022, li_model-contrastive_2021}. The details of the distributions are visualized in Figure~\ref{fig:data}. Specifically, the number of clients $N$, number of classes $C$, and the parameter of Dirichlet distribution $\alpha$ for each dataset are as follows:
\begin{itemize}
    \item CIFAR-100~\cite{krizhevsky_learning_2012}: $N=64, C=100,\alpha=0.1$.
    \item OrganAMNIST~\cite{medmnistv2}: $N=64, C=11, \alpha=0.5$.
    \item Office-Home~\cite{venkateswara2017deep}: $N=16, C=65, \alpha=1.0$.
\end{itemize}

\subsection{Hyperparameters}
We perform experiments based on the implementation from an existing
federated learning platform. For each method, we tune the hyperparameters in a range and report the result with the optimal hyperparameters. The range and optimal value of the hyperparameters are as follows:

    \textbf{FedAVG~\cite{mcmahan_communication-efficient_2017}}: Learning rate ($lr$) is searched from $\{0.001, 0.01, 0.1\}$. The optimal value is $0.01$.
    
    \textbf{Local}: $lr$ is searched from $\{0.001, 0.01, 0.1\}$. The optimal value is $0.01$.
    
    \textbf{APFL~\cite{deng_adaptive_2020}}: $lr$ and the initial mixture coefficient $\alpha$ are searched from $\{0.001, 0.01, 0.1\}$ and $\{0.25, 0.5, 0.75\}$, the optimal values are $lr=0.01$ and $\alpha=0.25$.

    \textbf{Per-FedAVG~\cite{fallah_personalized_2020}}: $lr$ and $\beta$ are searched from $\{0.001, 0.01, 0.1\}$ and $\{0.001, 0.01, 0.1\}$, the optimal values are $lr=0.001$ and $\beta=0.001$.

    \textbf{FedBN~\cite{li_fedbn_2021}}: $lr$ is searched from $\{0.001, 0.01, 0.1\}$. The optimal value is $0.01$.

    \textbf{FedRep~\cite{collins_exploiting_2021}}: $lr$ is searched from $\{0.001, 0.01, 0.1\}$. The optimal value is $0.01$. The classification head is defined as the last layer.

    \textbf{FedBABU~\cite{oh_fedbabu_2022}}: $lr$ is searched from $\{0.001, 0.01, 0.1\}$. The optimal value is $0.01$. One local step is done for fine-tuning the classification head. The classification head is defined as the last layer.

    \textbf{FedPerfix}: $lr$ is searched from $\{0.001, 0.01, 0.1\}$, and the optimal value is $0.01$. The hidden state dimension is set as $256$. The scale $s$ is set as $1.5$. The classification head is defined as the last two layers.

\section{Connection between APFL and FedPerfix}\label{sec:apfl}
In this section, we provide a detailed comparison between the APFL and FedPerfix to explain \textit{why they both have the leading performance} compared with other methods and show that our method has \textit{additional advantages in storage and computation resource requirements}. 

\subsection{Why APFL and FedPerfix lead the performance?}

First, we briefly introduce the idea of APFL. APFL keeps a separate personalized model for each client. In each communication round, it will first train the global and local models separately and obtain their gradients, then update the personalized model with the gradient mixed from these two models. Now if we only consider the personalized model, its update can be written as 
\begin{equation}
\begin{aligned}
 h_{per} &\leftarrow h_{per} - \eta( \alpha\nabla h_{per} + (1-\alpha) \nabla h_{global})\\
 &= h_{per} - \eta\nabla (\alpha h_{per} + (1-\alpha) h_{global}),
\end{aligned}
\end{equation}
where $h_{per}$ is the personalized model, $h_{global}$ is the global model, $\eta$ is the learning rate, and $\alpha$ is the mixture coefficient.

Therefore, the updating of the personalized model is equivalent to updating a model $\bar h$ that takes the mixture of the personalized and global models as the output. Further, we formulate the output of $\bar h$ as 
\begin{equation}
\label{eq:apfl}
\begin{aligned}
 \mO^{(L)}=&\textcolor{blue}{\alpha} h_{per}^{(L)}(\mZ^{(L-1)}) + \textcolor{blue}{(1-\alpha)} h_{global}^{(L)}(\mZ^{(L-1)}),
\end{aligned}
\end{equation}
where $L$ is the number of layers of the model, $\mZ^{(L-1)}$ is the hidden state from the last layer.

For FedPerfix, the output of one head of the self-attention layer can be formulated and rewritten~\cite{he_towards_2022} as 
\begin{equation}
\label{eq:prefix-supp}
\begin{aligned}
    head(\mZ) = &Attn(\mZ\mW_q, 
    \mZ[\mP_k,\mW_k], 
    \mZ[\mP_v,\mW_v])\\
    = & \text{softmax}(\mZ\mW_q[\mP_k,\mW_k]^{\top})\begin{bmatrix}
\mP_v\\
\mZ\mW_v
\end{bmatrix}\\
= & (1-\lambda(\mZ))\text{softmax}(\mZ\mW_q\mW_k^\top\mZ^\top)\mZ\mW_v 
\\& + \lambda(\mZ)\text{softmax}(\mZ\mW_q\mP_k^\top)\mP_v\\
= & \textcolor{blue}{(1-\lambda(\mZ))}Attn(\mZ\mW_q, 
x\mW_k, 
x\mW_v)\\
&+ \textcolor{blue}{\lambda(\mZ)}Attn(\mZ\mW_q, 
\mP_k, 
\mP_v)
\end{aligned}
\end{equation}
where $Attn$ is the attention operation, $Z$ is the hidden state from the last layer, and $\lambda(\mZ)$ is the mixture coefficient defined as
\begin{equation}
\small
\lambda(\mZ) = \frac{\sum_i\exp(\mZ\mW_q\mP_k^\top)_i}{\sum_i\exp(\mZ\mW_q\mP_k^\top)_i + \sum_j\exp(\mZ\mW_q\mW_k^\top\mZ^\top)_j}.
\end{equation}

If only taking the self-attention layer into consideration, Equation~\ref{eq:apfl} and Equation~\ref{eq:prefix-supp} share a similar formulation, which can be interpreted as an information transfer between the global and local models. Therefore, both approaches have a leading performance, indicating that \textit{the underlying idea to balance local and global information is effective and crucial in personalized federated learning}.

\subsection{Advantages of FedPerfix }
Although APFL and FedPerfix share a connected underlying idea, FedPerfix can outperform APFL in a consistent margin across different settings. Besides,
FedPerfix is much more efficient than APFL from several perspectives:
\begin{itemize}
    \item Parameter size of the Prefixes in FedPerfix (\textbf{289.0K}) is fewer than a separate self-attention layer (\textbf{577.5K}) in APFL, leading to fewer computational costs for the self-attention layer.
    \item APFL needs to perform the mixture between the global and local models with additional computational costs for every layer, while FedPerfix only needs to perform it for the self-attention layers.
    \item APFL needs to store a separate personalized model (\textbf{21.03M}) on each client, while FedPerfix only needs (\textbf{3.39M}) additional space.
\end{itemize}

In conclusion, APFL needs $\mathbf{70\times}$ additional FLOPs and $\mathbf{6.2\times}$ additional parameters to store than FedPerfix, using FedAVG as a baseline.
Therefore, when compared with APFL, FedPerfix not only achieves superior performance but also enjoys a more efficient implementation by specifying and focusing only on the sensitive parts of the ViT. This targeted approach leads to a more effective and efficient approach for personalized federated learning. 

\section{More Ablation Study}\label{sec:ab}
To reduce the exhausting hyperparameter-searching in practice, we report the result under a unified default setting for all tasks in the main paper. \textit{The performance under such a unified default setting still achieves state-of-the-art performance}, demonstrating the robustness of our proposed method.
However, we still want to demonstrate the potential to further improve the performance by tuning the hyperparameters. In this section, we will show the results and analyze the impact of three key hyperparameters in FedPerfix: hidden dimension, scale, and the depths of prefixes.
\subsection{Impact of Hidden Dimension and Scale}
The hidden dimension is the dimension of the hidden state of the adapter to generate the Prefixes, \ie, the common dimension shared by $\mW_{down}$ and $\mW_{up}$, and the scale $s$ is scalar to control the impact of the Prefixes. In our default setting, the hidden dimension is set as $256$, and the scale is set as $1.5$. FedPerfix, under the default setting, can outperform all the compared methods in all datasets. Here, we demonstrate the potential to achieve better performance on CIFAR-100 by tuning the hyperparameters.
As shown in Figure~\ref{fig:ablation} (a) and (b), increasing the hidden dimension and scale will not always lead to an increase in performance. Therefore, in practice, further tuning the hidden dimension and the scale can lead to a better result, even though without such tuning can still maintain a high performance.
\subsection{Impact of the Depths of Prefixes}
In our default setting, we add the Prefixes to every self-attention layer. To investigate the impact of the depths of Prefixes, we further conduct experiments only to add Prefixes to the last several self-attention layers, \ie, $depth=1$ means only adding Prefixes to the last self-attention layer. The result is shown in Figure~\ref{fig:ablation} (c). In general, with the increase in the depths of the Prefixes, the overall performance increases. However, such an increase is not strict, indicating the potential to further improve the performance and reduce storage and computational costs.

\end{document}